\documentclass[twoside,11pt]{article}

\usepackage{jmlr2e}

\usepackage{amsmath}
\usepackage{graphicx}
\usepackage{subfigure}
\usepackage{booktabs} 
\usepackage{upgreek}

\jmlrheading{}{2018}{}{}{}{University of Chicago}

\ShortHeadings{Aequitas: A Bias and Fairness Audit Toolkit}{Saleiro, Kuester, Stevens, Anisfeld, Hinkson, London and Ghani}
\firstpageno{1}

\begin{document}

\title{Aequitas: A Bias and Fairness Audit Toolkit}

\author{\name Pedro Saleiro \email pedrosaleiro@gmail.com \\
       \name Benedict Kuester  \\
       \name Loren Hinkson  \\
       \name Jesse London  \\
       \name Abby Stevens  \\
       \name Ari Anisfeld  \\
       \name Kit T. Rodolfa  \\
       \name Rayid Ghani \email rayid@uchicago.edu \\
       \addr Center for Data Science and Public Policy\\
       University of Chicago\\
       Chicago, IL 60637, USA}

\editor{}

\maketitle

\begin{abstract}

Recent work has raised concerns on the risk of unintended bias in AI systems being used nowadays that can affect individuals unfairly based on race, gender or religion, among other possible characteristics. While a lot of bias metrics and fairness definitions have been proposed in recent years, there is no consensus on which metric/definition should be used and there are very few available resources to operationalize them. Therefore, despite recent awareness, auditing for bias and fairness when developing and deploying AI systems is not yet a standard practice. We present Aequitas, an open source bias and fairness audit toolkit that was released in 2018 and it is an intuitive and easy to use addition to the machine learning workflow, enabling users to seamlessly test models for several bias and fairness metrics in relation to multiple population sub-groups. Aequitas facilitates informed and equitable decisions around developing and deploying algorithmic decision making systems for both data scientists, machine learning researchers and policymakers.

\end{abstract}


\section{Introduction}

Artificial Intelligence (AI) based systems are becoming ubiquitous in many different areas of practice in our society, from banking and insurance to human resource management, from criminal justice and law enforcement to healthcare. These systems are usually based on a Machine Learning (ML) model that is typically optimized for an utility measure focused on efficiency or effectiveness defined by the product managers and decision makers (e.g. AUC or overall accuracy). However, recent scrutiny of such systems such as the Gender Shades project  \cite{buolamwini2018gender} about how commercial facial recognition systems have disparate errors in detecting gender of darker skin women, or Google's efforts to reduce gender bias in machine translation \cite{kuczmarski2019translation}, has increased awareness around possible unintended biases that AI systems might have against people from specific groups, often under-represented, based on race, gender, religion or age, among other characteristics. 

While general purpose AI systems and their unintended biases might seem inoffensive when looked detached from specific applications, as the case of general purpose facial recognition systems or machine translation, these ML models might have dramatic consequences when applied to policy areas that have a direct impact in peoples lives such as the job market or surveillance and law enforcement, as demonstrated by the ProPublica article about COMPAS \cite{angwin2016machine} and subsequent discussions. 

These are daunting times. On one hand the rush for developing and deploying AI has resulted in remarkable achievements at an astonishing speed in both hardware and software for AI, on the other hand the development of new policies and practices adjusted to this new reality of ubiquitous AI, and its ethical implications, has been dramatically slower. 

The AI and Data Science research communities have been trying to respond to these concerns by 1) developing methods to \textit{detect} bias and discrimination in AI systems \cite{calders2010three,dwork2012fairness,zemel2013learning,hardt2016equality,zafar2017parity,zafar2017fairnessw}, and 2) developing methods to \textit{mitigate} bias and disparate impact or defining tradeoffs among different criteria \cite{hardt2016equality,kamishima2011fairness,feldman2015certifying,kleinberg2016inherent,corbett2017algorithmic,zafar2017fairnessw, kearns2017preventing, noriega2018active}. A lot of this work has focused on a single disparity measure between only two groups (e.g. female vs male, black vs white), analyzing tradeoffs for a pair of measures, using synthetic datasets, toy UCI data sets, or applied to a single problem. 

There has been very little extensive empirical work done on computing and evaluating a wide variety of various bias metrics and fairness definitions on real world public policy problems. There is also no consensus on what happens across policy problems in different domains in practice, and what tradeoffs and possible solutions should be used. Moreover, a recent comparative study \cite{friedler2018comparative} of bias mitigation methods has shown that 
fairness interventions might be more brittle than previously thought. We believe that in order for AI to have a beneficial policy impact, the research community has to do more on the applied side of this problem, evaluate and operationalize our methods on real policy problems, and provide tools for data scientists and policymakers to help them make informed decisions. To overcome these barriers we developed Aequitas, an open source\footnote{\url{https://github.com/dssg/aequitas}} bias and fairness audit toolkit that was released in May 2018 \footnote{\url{https://twitter.com/datascifellows/status/994204100542783488}} and it is an intuitive and easy to use addition to the ML workflow, enabling users to seamlessly audit ML models for several bias and fairness metrics in relation to multiple population sub-groups. Aequitas can be used directly as a Python library, via command line interface or a web application, making it accessible and friendly to a wide range of users (from data scientists to policymakers). 

We believe that systematic audits for bias and fairness is the first step to make informed model selection decisions, to better understand the causes of bias against specific groups, and to create trust in AI among the general population.  Aequitas contributes to make auditing for bias and fairness a standard procedure when developing, maintaining or considering deploying AI systems. Aequitas had an imediate impact in both the data science and public policy communities and it was featured in a Nature\footnote{\url{https://www.nature.com/articles/d41586-018-05469-3}} news article about ``bias detectives''. The web application\footnote{\url{http://aequitas.dssg.io/}} alone has served more than 2400 sessions from 1300 unique users, while the source code had an average of 300 views and 20 clones per week.


\section{Related Work}

Fairness has been defined has being being accountable on the treatment \cite{pedreshi2008discrimination,dwork2012fairness,luong2011k,calders2013unbiased,hardt2016equality} or impact level\cite{kleinberg2016inherent,chouldechova2017fair,feldman2015certifying}. There is fairness on the treatment when the sensitive attributes are not explicitly used in the modeling. Fairness on impact is concerned only with the predictions and can be measured using ground-truth label information or not. This work focuses on assessing fairness on the impact using parity based measures.

Impact is a very common notion of fairness in previous work \cite{calders2010three,hardt2016equality,zafar2017fairnessw,feldman2015certifying,zemel2013learning,kamishima2011fairness,zafar2017fairness}. A predictor has parity on impact when the fraction of elements of the group predicted as positive is the same across groups. On the other hand, a predictor has statistical or demographic parity \cite{calders2009building,luong2011k,zemel2013learning,dwork2012fairness,zliobaite2015relation,zafar2015learning,corbett2017algorithmic} if there is an equal fraction of elements of each group among all predicted positives, i.e., the ``benefits'' are equally distributed.

In statistical risk assessment a risk score is considered calibrated, or test-fair \cite{kleinberg2016inherent,chouldechova2017fair} if it has equal precision among different groups (e.g. male vs female) for every value of the predicted risk score. This is a notion similar to equalized odds \cite{hardt2016equality}, which consists in having equalized false negative rate and false positive rates. When the application gives more importance to the positive outcome (``advantage''outcome), Hardt et al.\cite{hardt2016equality} propose the notion of equal opportunity, which consists in relaxing the equalized odds notion to just care for the false negative rate parity. 

Kleinberg et al.\cite{kleinberg2016inherent}, Joseph et al. \cite{joseph2016fairness} and Chouldechova \cite{chouldechova2017fair} discuss the relation between model calibration, prevalence, false negative and false positive rates in risk assessment prediction. Prevalence corresponds to the fraction of data points labeled as positive on each group. Both authors show that when there is calibration and the prevalence is different between groups then it is not possible to have both equal false positive and negative rates across groups, i.e., balance for the positive and negative classes. 

There is also the notion of individual fairness, which states that similar entities should receive similar predictions regardless of the sensitive attribute. Dwork et al. uses a distance measure between entities and tries to model classifiers having similar predictions among similar entities, using a Lipschitz condition. Luong et al. \cite{luong2011k} propose a similar notion but focused on entities on similarity of entities having a negative outcome.

Previous work in this line of research has not focused on empirical analysis of the definitions of bias neither empirical validation of the proposed solutions, being very common the use of synthetic datasets, with exception of the use of COMPAS, for obvious reasons. Recent work of Chouldechova et al. \cite{chouldechova2018case} presents a case study of using a risk assessment model to predict adverse outcomes based on child maltreatment referrals in the state of Pennsylvania. 

In the past couple of years there have been released many github repositories dedicated to bias and fairness in ML. The majority of these repositories comprise ad-hoc analyses and code implementations of the bias and fairness definitions proposed in the literature. For example, FairTest \cite{tramer2017fairtest} focused in the upstream detection of bias by analyzing training sets and checking for correlations between labels and protected attributes. FairML \cite{adebayo2016} is a library that can be categorized as an attempt to increase interpretability of the predictions of black box models with respect to perturbations in the inputs, and therefore can also be used to measure how different models react to data points representing different sub-groups. Fairness Measures \cite{zehlike2017} provides several fairness measures and it is focused on learning to rank problems. It is followed by a website containing the descriptions of the measures implemented in the repository. Recently, IBM released AIFairness360 \cite{bellamy2018ai}, a general purpose fairness toolbox that compiles and implements both detection and the state-of-the-art methods for bias mitigation. 

Aequitas distinguishes from the existing open source repositories and toolboxes in two key aspects. First, bias and fairness are not absolute concepts and are not independent from the application scenario, as well as its analysis and interpretation. Aequitas provides comprehensive information on how it should be used in a public policy context, taking the resulting interventions and its implications into consideration. Second, Aequitas is intended to be used not just by data scientists but also policymakers, and consequently provides seamless integration in the ML workflow but also provides a web app tailored for non-technical people auditing ML models.

\section{Aequitas}

Aequitas can audit AI systems to look for biased actions or outcomes that are based on false or skewed assumptions about various demographic groups. Using a Python library a command line interface, users simply upload data from the system being audited, configure bias metrics for protected attribute groups of interest as well as reference groups, and then the tool generates bias reports. 

\begin{figure}[t]
\centering
\includegraphics[width=0.8\columnwidth]{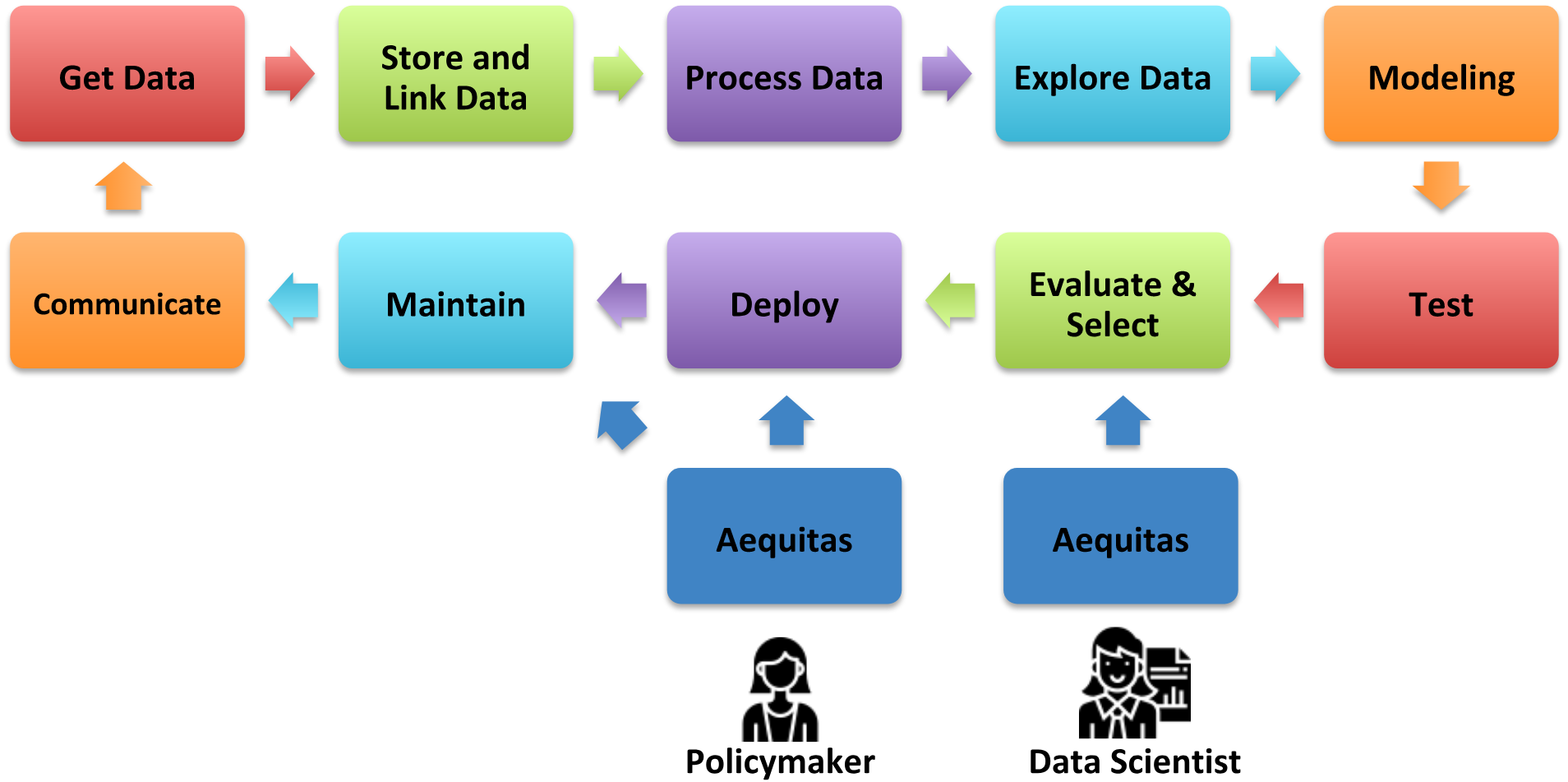}
\caption{Aequitas in the larger context of the ML pipeline. Audits must be carried internally by data scientists before evaluation and model selection. Policymakers (or clients) must audit externally before accepting a model in production as well as perform periodic audits to detect any fairness degradation over time.}\label{fig:auditpipeline}
\end{figure}

\subsection{The Audit Paradigm}

In Aequitas, bias assessments can be made prior to model selection, evaluating the disparities of the various models trained based on whatever training data was used to tune it for its task. The audits can be performed prior to a model being operationalized, based on operational data of how biased the model proved to be in holdout data. Or they can involve a bit of both, auditing bias in an A/B testing environment in which limited trials of revised algorithms are evaluated whatever biases were observed in those same systems in prior production deployments.

Aequitas was designed to be used by two types of users:
\begin{enumerate}
\item Data Scientists and AI Researchers: who are building AI systems for use in risk assessment tools. They will use Aequitas to compare bias measures and check for disparities in different models they are building during the process of model building and selection.
\item Policymakers: who, before ``accepting'' an AI system to use in policy decision, will run Aequitas to understand what biases exist in the system and what (if anything) they need to do in order to mitigate those biases. This process must be carried periodically to assess the fairness degradation through time of a model in production.
\end{enumerate}

Figure \ref{fig:auditpipeline} puts Aequitas in the context of the ML workflow and shows which type of user and when the audits must be carried. The main goal of Aequitas is to make standard the audit paradigm, i.e., by auditing both internally (data scientists) and externally (policymakers) it forces the different actors to have bias and fairness always into consideration when making decisions around model selection, deployment or not, the need to retrain, the need to collect more and better data, and so on.

%

\subsection{Measuring Bias and Fairness}

A large number of public policy and social good problems where AI systems are used to help human experts make decisions share some common characteristics, including skewed class distribution, and the metric of interest being precision at top $k$, as opposed to accuracy or AUC-ROC in many other machine learning problems. This is the case because we often have limited intervention resources, and can only take action on a small number ($k$) of entities. 

\begin{figure}[h]
\centering
\includegraphics[width=0.6\columnwidth]{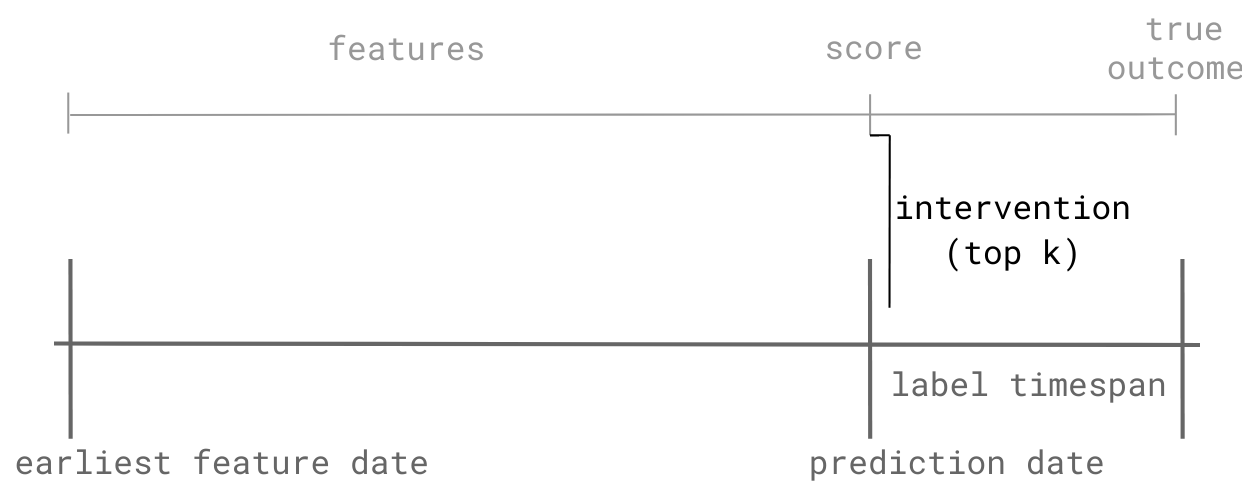}
\caption{Algorithmic Decision Making timeline for Public Policy and Social Good problems.}
\label{fig:policy-prediction}
\end{figure}

The goal of the ML model is to accurately prioritize the top $k$, which is equivalent to maximizing the precision at top $k$. Figure \ref{fig:policy-prediction} depicts a common timeline of AI systems for policy and social good. Based on the risk score (prediction) produced by the ML model the entities are ranked, and often with a human expert in the loop, the top $k$ entities are selected to intervene on. These interventions can either be assistive (helping individuals with housing assistance to reduce their risk of future homelessness), or sometimes punitive (housing inspections that would result in fines and repair costs if violations are found). 

A traditional binary classification task using supervised learning consists of learning a predictor $\widehat{Y} \in \{0,1\}$, that aims to predict the true outcome $Y \in \{0,1\}$ of a given data point from the set of features $X$, based on labeled training data. Many problems in public policy can be formulated as statistical risk assessment problems in which we assign a real valued score $S \in [0,1]$ to each entity (data point) and a decision $\widehat{Y}$ is made based on the score, typically by selecting a pre-defined number (k) of entities that should be classified as positive. After sorting the entities by $S$, the binary predictor is defined as  $\widehat{Y}=1$ if $R \geq s_k$ where $s_k$ is the score of the kth ordered entity.
The main definitions of this sub-section are the following:

\textbf{Score} - $S \in [0,1]$ is a real valued score assigned to each entity by the predictor. 

\textbf{Decision} - $\widehat{Y} \in \{0,1\}$ is a binary prediction assigned to a given entity (data point), based on thresholding on the score (e.g. top \textit{K}).                       

\textbf{True Outcome} - $Y \in \{0,1\}$ is the true binary label of a given entity.          

\subsubsection{Defining Groups}

Let us now consider a multi-valued attribute $A = \{a_i, a_2,...,a_n\}$ that can be or not be a subset of $X$, for example gender$=\{$\textit{female}, \textit{male}, \textit{other}$\}$. We define a group $g(a_i)$ as a set of entities (data points) that have in common a specific attribute value of $A=a_i$, for instance gender=\textit{female} corresponding to all the females in the dataset. Given all groups defined by the attribute $A$, the predictions $\widehat{Y}$ and true outcome $Y$ for every entity of each group we can now discuss group metrics. 
The main definitions about defining groups for assessing bias and fairness are the following:

\textbf{Attribute} - $A = \{a_i, a_2,...,a_n\}$ is a multi-valued attribute, e.g., gender$=\{$\textit{female}, \textit{male}, \textit{other}$\}$    

\textbf{Group} - $g(a_i)$ is a group of all entities that share the same attribute value, e.g., gender=female.

\textbf{Reference Group} - $g(a_r)$ is one of the groups of A that is used as reference for calculating bias measures.

\textbf{Labeled Positive} - $LP_g$ is the number of entities labeled as positive within a group.                       

\textbf{Labeled Negative} - $LN_g$ is the number of entities labeled as negative within a group.                       

\textbf{Prevalence} -$Prev_g$ = $LP_g$\ \textbf{/}\ $|g| = \text{Pr(Y=1|A=}a_i)$ is the fraction of entities within a group which true outcome was positive.    \\

\subsubsection{Distributional Group Metrics}
We can now define decision making metrics at group level. We use two metrics (Predicted Prevalence and Predicted Positive Rate) that are only concerned about the distribution of the entities across groups in the selected set for intervention (top k) and therefore do not use the true outcomes (labels). We define distributional group metrics as follows: 

\textbf{Predicted Positive}      - $PP_g$ is the number of entities within a group where the decision is positive,i.e.,  $\widehat{Y}=1$. 

\textbf{Total Predictive Positive} - K = $\sum_{A=a_1}^{A=a_n}$ $PP_{g(a_i)}$ is the total number of entities predicted positive across groups defined by $A$.                                

\textbf{Predicted Negative}      - $PN_g$ is the number of entities within a group which decision is negative,i.e.,  $\widehat{Y}=0$.   

\textbf{Predicted Prevalence}    -$PPrev_g$ = $PP_g$\ \textbf{/}\ $|g| = \text{Pr(}\widehat{Y}\text{=1|A=}a_i)$ is the fraction of entities within a group which were predicted as positive.                              \

\textbf{Predicted Positive Rate} - $PPR_g$ = $PP_g$\ \textbf{/}\ $K = \text{Pr(a=}a_i|\widehat{Y}\text{=1)} $ is the fraction of the entities predicted as positive that belong to a certain group.    \\

\subsubsection{Error-based Group Metrics}
We now define group metrics that require the true outcome (label) to be calculated. We focus on type I (false positives) and type II (false negative) errors across different groups. In the context of public policy and social good the goal is to avoid disproportionate errors in specific groups. We use four different error-based group metrics defined as follows:

\textbf{False Positive}          - $FP_g$  is the number of entities of the group with $\widehat{Y}=1 \land Y=0$ 

\textbf{False Negative}          - $FN_g$ is the number of entities of the group with $\widehat{Y}=0 \land Y=1$. 

\textbf{True Positive}           - $TP_g$ is the number of entities of the group with  $\widehat{Y}=1 \land Y=1$. 

\textbf{True Negative}           - $TN_g$ is the number of entities of the group with  $\widehat{Y}=0 \land Y=0$.

\textbf{False Discovery Rate}    - $FDR_g$ = $FP_g$\ \textbf{/}\ $PP_g = \text{Pr(Y=0|}\widehat{Y}\text{=1,A=}a_i)$ is the fraction of false positives of a group within the predicted positive of the group

\textbf{False Omission Rate}     - $FOR_g$ = $FN_g$\ \textbf{/}\ $PN_g = \text{Pr(Y=1|}\widehat{Y}\text{=0,A=}a_i)$ is the fraction of false negatives of a group within the predicted negative of the group

\textbf{False Positive Rate}     - $FPR_g$ = $FP_g$\ \textbf{/}\ $LN_g  = \text{Pr(}\widehat{Y}\text{=1|Y=0,A=}a_i)$ is the fraction of false positives of a group within the labeled negative of the group

\textbf{False Negative Rate}     - $FNR_g$ = $FN_g$\ \textbf{/}\ $LP_g = \text{Pr(}\widehat{Y}\text{=0|Y=1,A=}a_i)$ is the fraction of false negatives of a group within the labeled positives of the group

\subsubsection{Assessing Group Bias and Fairness on Impact}
In the context of public policy and social good we want to avoid providing less benefits to specific groups of entities, if the intervention is assistive, as well as, avoid hurting more specific groups, if the intervention is punitive. Therefore we define bias as a disparity measure of group metric values of a given group when compared with a reference group. This reference can be selected using different criteria. For instance, one could use the majority group (with larger size) across the groups defined by $A$, or the group with minimum group metric value, or the traditional approach of fixing a historically favored group (e.g race:white).  The bias measures are applied on a pairwise basis comparing groups defined by a given attribute $A=a_i$. For instance we define Predicted Prevalence Disparity as
\begin{equation}
PPRev_g \ disp = \frac{PPrev_{a_i}}{PPrev_{a_r}} = \frac{\text{Pr(}\widehat{Y}\text{=1|A=}a_i)}{\text{Pr(}\widehat{Y}\text{=1|A=}a_r)}
\end{equation}

and False Positive Rate Disparity as: 

\begin{equation}
FPR_g \ disp = \frac{FPR_{a_i}}{FPR_{a_r}} = \frac{\text{Pr(}\widehat{Y}\text{=1 | Y=0,A=}a_i) }{\text{Pr(}\widehat{Y}\text{=1 | Y=0,A=}a_r) }
\end{equation}

\
We use parity based measures of impact fairness. Our formulation and implementation of fairness is flexible as it relies on a real valued parameter $\uptau \in (0,1] $ to control the range of disparity values that can be considered fair. One example, of formulation of disparity is using the ```80\% rule'' represented by $\uptau = 0.8$. A predictor must be as fair as the maximum value of the bias across the groups defined by $A$ allow. This notion of \textbf{parity} requires that all biases to be within the range defined by $\uptau$:

\begin{equation}
\tau \leq Disparity Measure_{group_i} \leq \frac{1}{\tau}
\end{equation}

The different fairness measures vary in importance to the end-user based on the cost and impact of the intervention. If the interventions are very expensive or could hurt the individuals, then we would want to minimize false positives (focusing on the False Discovery Rate and/or the False Positive Rate). If the interventions are predominantly assistive, we should be more concerned with false negatives (focusing on the False Omission Rate and/or the False Negative Rate). 

\subsection{Gaps and Barriers}

The main barrier we have witnessed in the adoption of audits for bias and fairness has been in creating a systematic and sustained interaction and efforts between the policymakers and the data scientists/AI researchers who develop the AI systems. AI and data science are relatively new fields where there is no established tradition in studying the ethical consequences of the usage of biased models. At the same time, policymakers have little experience in dealing with AI and are not technically equipped to make specific design decisions. Consequently, although the need for audit bias and fairness might sound obvious, it is not a standard practice nowadays. Aequitas is a step forward in that direction.

\begin{figure}[h]
	\centering
	\includegraphics[width=1.0\linewidth]{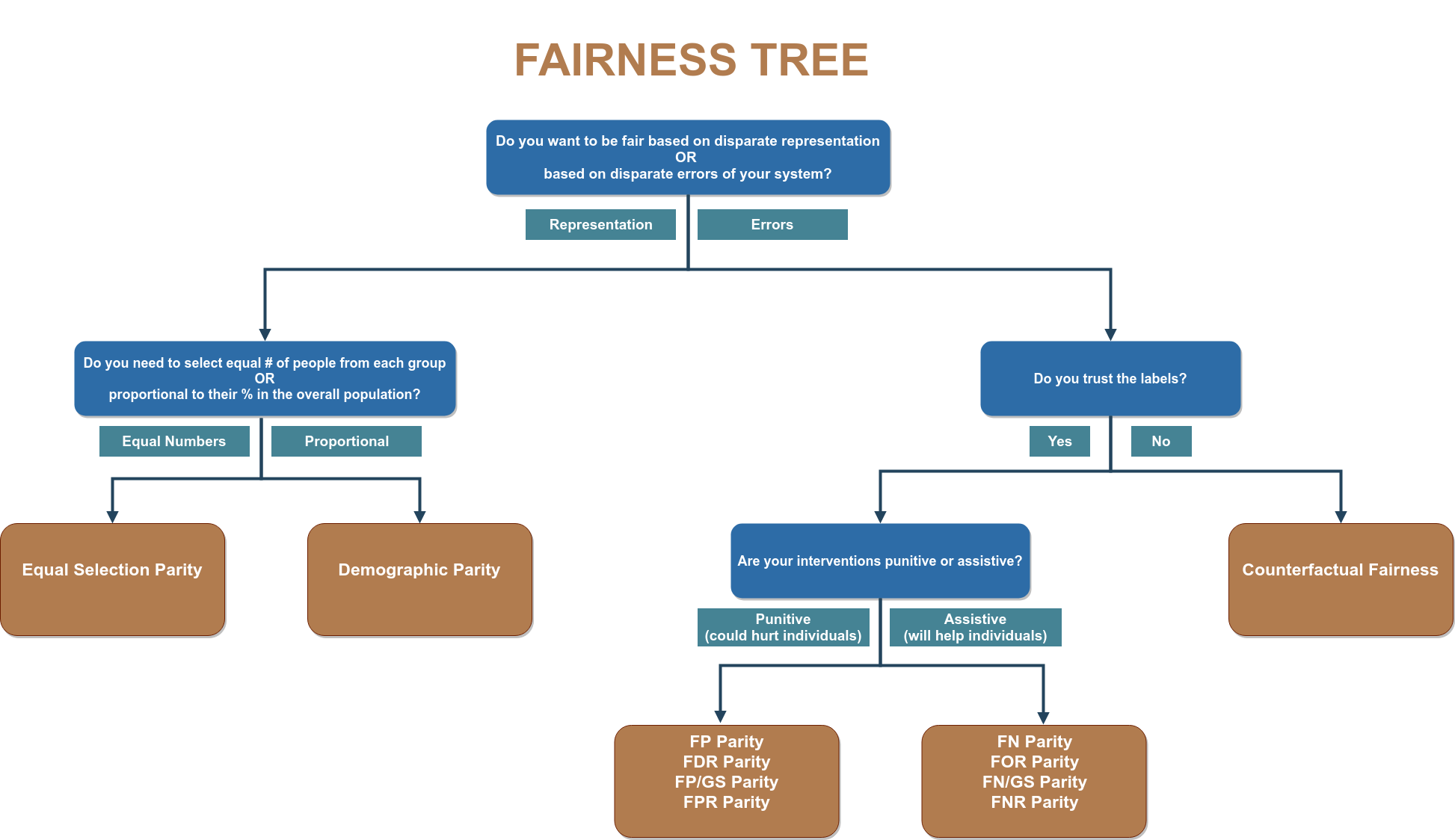}
	\caption{Fairness tree helps both data scientists and policymakers to select the fairness metric(s) that are relevant to each context.}
	\label{fig:fairness-tree}
\end{figure}

Before releasing Aequitas, we asked both data scientists and policymakers to use the toolkit in real-world projects to collect feedback. Interestingly, both types of users complained that it was hard to navigate in so many different metrics and there was no guidance in how to link the different metrics and the real-world problem in hand. Therefore, we designed the ``Fairness Tree'' (Fig. \ref{fig:fairness-tree}) in collaboration with policymakers, which represent a complete navigation of the most relevant bias metrics implemented in Aequitas. It is designed from the decision maker perspective and assume that there are fundamental policy options that the decision maker has decided in the beginning of a data science/AI project. For instance, the tree asks if the decision maker will carry on interventions based on the model predicted labels, and if those interventions will help people or if they will hurt people. This is a friendly way of asking the user if he cares more about disparate distribution of false negatives or false positives.

\section{Case Studies}

To show the impact of using Aequitas, we build on the Center for Data Science and Public Policy experience on large scale data driven problems in public health, criminal justice and public safety. We present an extensive empirical comparison of several bias and fairness measures across different real-world public policy projects where algorithmic decision making systems were trained and used to predict risk scores. To the best of our knowledge this is the most extensive public disclosure of bias audits performed in real-world ML models being used in different policy areas.

\subsection{Criminal Justice}

With millions of people moving through courts and jails every year in the US, the criminal justice system is larger than ever. Communities across the country have recognized that a relatively small number of these highly vulnerable people cycle repeatedly not just through criminal justice system, but also emergency medical services (EMS), hospital emergency rooms, homeless shelters, and other public systems, receiving fragmented and uncoordinated care with poor outcomes at great cost. This project with a large US city (referenced redacted for blind review) was aimed at identifying individuals who repeatedly cycle through the criminal justice system and help the city determine tailored, preventative interventions for these individuals to reduce their risk of recidivism.

\begin{itemize}
\item \textbf{Goal}: The specific goal was to use historical data about individuals, predict their likelihood of recidivism in the next 6 months, and match the 150 highest risk individuals with tailored, preventative interventions.
\item \textbf{Data}: The data used was criminal justice data for 1.5 Million individuals over the past 10 years, primarily consisting of cases and bookings. The analysis focused on around 400,000 individuals who had repeated interactions with the criminal justice system.
\item \textbf{Performance Metric}: Since the pilot would involve intervening with 150 individuals, we used Precision at the top 150 as our performance metric to optimize.
\item \textbf{Analysis}: The analysis involved integrating data from multiple sources using record linkage techniques, generating over 3000 features (ranging from demographic to behavioral to spatio-temporal aggregations over different time and space resolutions) and building several hundred models. We  validate our  models  using  temporal cross-validation \cite{hyndman_forecasting:_2014}. We split the data into several train-test splits by time, fitting models on all criminal justice interactions occurring before year $t$ and testing on data occurring in year $t$ + 6 months. (The 6 month gap between train and test data is due to the outcome window, which can extend 6 months after the last day of year $t-1$.) We train and test repeatedly by moving 6 months forward.
Based on this model selection process, we select a model that optimizes our metric -- with 73\% precision at the top 150. We used two baselines to compare it against: 1) random baseline with 4.4\% precision and 2) the existing heuristics used by the city today using past frequency of criminal justice interactions to prioritize individuals, with 43\% precision.
\end{itemize}

We now audit the selected model that is ready for deployment for bias.

\begin{figure*}[h]
\centering
\includegraphics[width=1.0\linewidth]{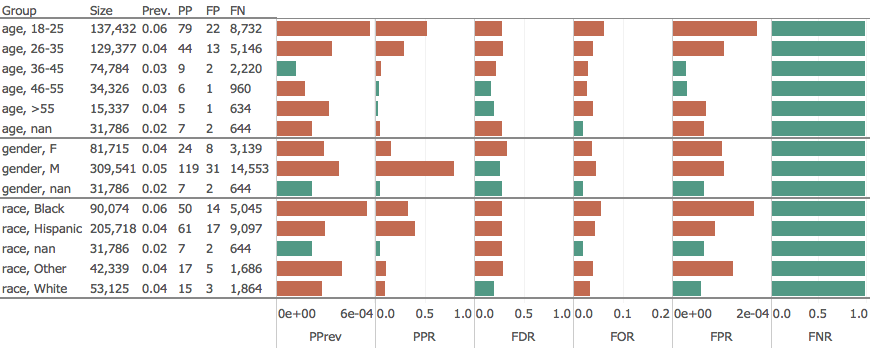}
\caption{Recidivism, evaluation results per group value for selected model (Precision@150 = 0.73).}
\label{fig:cj-detail}
\end{figure*}

We find that there is indeed unfairness in the model. Going one level down, we find that there is both supervised and unsupervised unfairness in all three attributes of interest: age, gender, and race. 

Figure \ref{fig:cj-detail} shows the detailed audit results. Each row represents a specific attribute-value pair (gender:female), and each bar (column) represents a group metric of interest (False Positive Rate for example).
Green bars represent groups for which the model does not exhibit bias within that metric. Red bars are those that are unfavorably biased compared to the reference group.

Although the tool produces results for all bias measures, since the interventions we're focusing on in this problem are ``assistive'', we only need to consider Type II parity -- False Omission Rates (FOR) and False Negative Rates (FNR). This is because our interventions are preventative and designed to provide extra assistance to individuals. Providing this assistance to individuals who are false positives will not hurt them but missing individuals could be harmful to them. To make sure that we don't disproportionately miss individuals from groups of interest, we want to focus on Type II parity.

Further, in this case, we are looking at prioritizing 150 individuals from a set of around 400,000, false negative rate parity is not appropriate to use either, leaving us to focus on False Omission Rate parity. Looking at Figure \ref{fig:cj-detail}, we find that the selected model:

\begin{itemize}
\item is fair for all values of Gender.
\item Young Adults (ages 18-25) have FOR disparity of 2X, having FOR of 6\% compared to the group with the lowest FOR (36-55 year olds), that have FOR of 3\%. 
\item Individuals with ages 26-35 or 55+ also have FOR disparity of 1.5X.
\item African Americans have FOR disparity of 1.6x , having an FOR of 5.6\% compared to the reference group (White) having FOR of 3.5\%.
\item Hispanics have FOR disparity of 1.3x compared to the reference group (whites) having FOR equal 4.4\% compared to 3.5\%.
\end{itemize}

Now, we turn to looking at the unsupervised fairness results (the first two columns of bars in Figure \ref{fig:cj-detail}).

Looking at statistical parity (second column labeled PPR), we find that, in the top 150 individuals scored, there are:
\begin{itemize}
\item more Males (79\%) than Females (16\%).
\item more Young people (82\% under 35) than Older people.
\item more African American (33\%) and Hispanics (41\%) than White (10\%).
\end{itemize}

Looking at Impact parity (first column labeled PPrev), we find that, in the top 150 individuals scored, there is Disparate Impact of :\begin{itemize}
\item 5X for Young People (ages 18-35). 0.06\% of them are in the top 150 compared to 0.012\% of people aged 36-45.  
\item 2X for African Americans. 0.055\% of them are in the top 150 compared to 0.028\% for White.  
\end{itemize}

Looking at these bias audit results, it seems that there are some biases present in our model for metrics that we care about. To compare these biases with the current policy being used by the city, we ran the same analysis again. We don't show the detailed results because of space constraints. The existing approach uses the recent frequency of interactions with the criminal justice system to prioritized 150 individuals. Analyzing the bias results, we find that that approach is more biased, not only for the FOR parity, but also for FDR parity.

\subsection{Public Health}

Retaining individuals living with HIV in care has been shown to be critical in reducing onwards HIV transmission. Individuals retained in care are less likely to develop AIDS and transmit the infection to others and are more likely to have longer lifespans than their unretained counterparts; yet maintaining quarterly appointments and daily medication for a lifetime is exceedingly difficult. 
This project focuses on the problem of prioritizing individuals for retention interventions based on their likelihood of dropping out of care. In partnership with an HIV clinic and the Department of Public Health of a major US city (referenced redacted for blind review), we have developed a predictive model that provides a risk score for whether a patient will be retained in care at the time of the patient's doctor visit. This risk score and additional information extracted from the machine learning model is then used by the clinic to assign personalized interventions to the patient to increase their retention likelihood. 

\begin{itemize}
\item \textbf{Goal}: The goal of this work was to build a point-of-service machine learning system that, at the time of a clinical visit, assesses the HIV patient's risk of not returning for continued treatment, as well as the associated risk factors. The definition of retention used is the patient having at least two appointment 90 days apart within a 12 month time period. 

\item \textbf{Data}: The primary source of data used in the model are the electronic health records of all the patients receiving care from the HIV Clinic. The patient records span 8 years from 2008 - 2016 and include approximately 1,600 patients. This is a rich data source that includes data on all appointments in the health system, hospital encounters, infectious disease specialist visits, visits with other types of physicians, co-morbidities, list of prescribed drugs, opportunistic infections, substance abuse data, demographic data, location data, lab tests, and doctor's notes. 
\item \textbf{Metric}: The HIV Clinic has an estimated capacity to provide interventions for 10\% of their patients. Thus, we selected models to maximize precision on the top 10\% of at-risk patient visits. 
\item \textbf{Analysis}: We augment the medical records from HIV Clinic with additional information such as environmental data and crime records. We generate 1600 features from this augmented data set and tested a number of algorithms, including Decision Trees, Random Forests, Logistic Regression, and Gradient Boosting, with various combinations of hyperparameters. 

For each time period, and each model, we calculate the difference in performance (based on our metric -- precision at the top 10\%) of that model with that of the model that performed the best in that time period. We then take all the previous time periods (from the date of deployment), and select the model whose performance was most consistently closest to that of the best model over all time periods. We found that shallow Random Forests with a maximum depth of 1 and 10,000 estimators tended to work well on this problem.

We compare the performance of the machine learning models to two baselines:
\begin{enumerate}
\item Random Baseline: Based on the class priors (fraction of appointments with no follow-up -- approximately 10\%)
\item Clinician Generated Rules: This expert baseline is meant to mimic an expert (health practitioner) making the prediction of whether the patient will remain in care or not.

We find that our machine learning model improves over the expert heuristic model by 20\%-50\%.
\end{enumerate}
\end{itemize}

Since this project deals with extremely vulnerable populations in our society, we run a series of bias audits across different bias measures for sensitive attributes of the patients.

Because the interventions we're focusing on here are ''assistive'' interventions (as opposed to punitive interventions), we consider parity in False Omission Rates (FOR) and False Negative Rates (FNR) for the groups we consider ''protected''.

\begin{figure*}[h]
\centering
\includegraphics[width=1.0\linewidth]{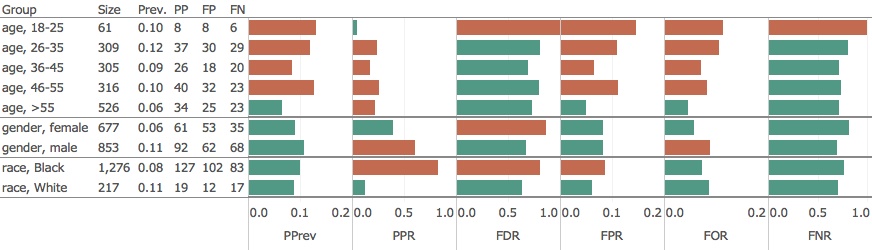}
\caption{HIV project, evaluation results  per group value for selected model (Precision@100 = 0.24).}
\label{fig:hiv}
\end{figure*}

We compare the bias results of our best performing model (Random Forest) with the expert (baseline) model. We only show results for the Random Forest in Figure \ref{fig:hiv} due to space constraints.

In both models, females (FOR-Expert = 6.4\%, FOR-RandomForest = 5.7\%) are favorably biased in terms of FOR than males (FOR-Expert = 10\%, FOR-RandomForest = 8.9\%) -- they are less often incorrectly missed.   In fact, for both groups, the Random Forest is less biased than the expert model. 

For race, white individuals (FOR-Expert = 10.7\%, FOR-RandomForest = 8.6\%) have a higher FOR than African-Americans (FOR-Expert = 7.7\%, FOR-RandomForest = 7.2\%), and are more often misclassified as false negatives. The expert model makes this mistake to a greater degree compared to the Random Forest for both groups.

Another group with high relative FNR is young adults in the age range of 18-25. This is especially crucial to be aware of since in the target city, this is a high-risk group. 

This in-depth bias audit analysis shows that while there are some possible biases in the model we've developed, these biases are much more pronounced in the expert model that is the current approach. We plan to work with the public health policymakers to further understand these biases and make an informed policy decision moving forward before deploying this model.

\subsection{Public Safety and Policing}
Recent high-profile cases of police officers using deadly force against members of the public have caused a political and public uproar. Adverse events between the police and the public thus come in many different forms, from deadly use of a weapon to a lack of courtesy paid to
a victim’s family. This project was focused on building a data-driven Early Intervention System that would identify police officers at risk of an adverse event in the near future. The police department would use the risk scores to identify officers in need of interventions (additional training, counseling, or other programs). Here, we focus on the models that were built and deployed at one of the police departments we worked with, in a major US city. More details on this project are available in (referenced redacted for blind review).

\begin{itemize}
\item \textbf{Goal}:  Given the set of all active officers at a given date and all data collected by a police department prior to that date, predict which officers will have an adverse interaction in the next year.
\item \textbf{Data}: The data for this work consists of almost all employee information and event records collected by PD to manage its day-to-day operations. This included data from Internal Affairs, Dispatch Events, Criminal Complaints, Citations, Traffic Stops, Arrests, Field Interviews, Employee Records, Secondary Employment, and Training events over the past 15 years, totaling over 20M records.
\item \textbf{Performance Metric}: The department had intervention resources to focus on 10\% of the police force. Thus, we use precision at top 10\% as our metric to optimize.
\item \textbf{Analysis}: We generated over 400 features and tested various models including Random Forests, Support Vector Machines, Logistic Regression, Gradient Boosted Trees. We used temporal cross-validation to perform model selection (similar to what we described in Case Study 1) from 2009-2016. Our best performing model is able to flag $10-15\%$ more high-risk officers (true positives), while reducing false positive rate by  $40\%$ compared to the current Early Intervention System being used by PD. The selected model was then put into a pilot in 2016 to validate predictions going forward, and based on the successful pilot, we deployed this system at PD in November 2017.

\end{itemize}

\begin{figure*}[t]
\centering
\includegraphics[width=1.0\linewidth]{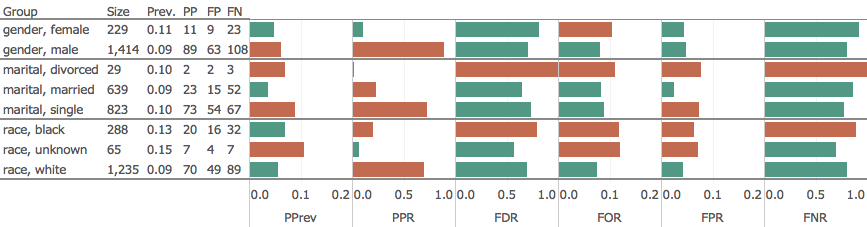}
\caption{Police Early Intervation System, evaluation results per group value for selected model (Precision@100 = 0.28).}
\label{fig:cmpd}
\end{figure*}

We now audit the selected model that was deployed  for biases using our Bias Audit Tool. At a high level, we find that there is indeed bias in the model. Going one level down, we find that there is bias in all three attributes of interest: gender, race, and marital status. In this problem, we're interested in both Type 1 and Type II parity, and in FDR parity and FOR parity specifically. 

Looking at Figure \ref{fig:cmpd}, we find that the selected model:

\begin{itemize}
\item is unfavorably biased for Divorced officers, and to a lesser extend, for African American officers in terms of FDR parity. 
\item is slightly unfavorably biased for Females, Divorced, and African American and Unknown Race officer in terms of FOR parity.

\end{itemize}

Now, we turn to looking at the unsupervised fairness results (the first two columns of bars in Figure \ref{fig:cmpd}).

Looking at statistical parity (second column labeled PPR), we find that, in the top 150 individuals scored, there are:
\begin{itemize}
\item more Males than Females.
\item more Single officers than Divorced or Married. 
\item more White officers than African-American or unknown race.
\end{itemize}

Looking at Impact parity (first column labeled PPrev), we find that, in the top 150 individuals scored, there is Disparate Impact for :\begin{itemize}
\item 5X for Young People (ages 18-35). 0.06\% of them are in the top 150 compared to 0.012\% of people aged 36-45.  
\item 2X for African Americans. 0.055\% of them are in the top 150 compared to 0.028\% for White.  
\end{itemize}


\section{Conclusions and Future Work}
AI systems are being increasingly used in problems that can have drastic impact on people's lives in policy areas such as criminal justice, education, public health, workforce development and social services. In this paper, we presented Aequitas, a toolkit for auditing group bias and fairness that is tailored to help AI developers and policymakers audit the output of AI systems for decision making in Public Policy problems. Auditing bias and fairness before accepting an AI system to production might seem a very obvious practice, however it is not yet a standard procedure. Aequitas makes a strong contribution in bringing together data scientists and policymakers to make more equitable decisions around developing, deploying and maintaining ML models in the real-world.

We used Aequitas to perform audit and fairness audit on real-world problems from three different policy areas: criminal justice, public health and public safety and policing. Aequitas allowed us to find that many of the ML models deployed for these problems do indeed have biases, but in most of those cases, the alternatives being used by policymakers today, are much more biased. The AI systems tend to be more accurate, and either equally or less biased, in effect improving equity and fairness of the policy. These results suggest that well-audited machine learning models are more effective at both solving the policy problem as well as reducing inequities. 

The work here is a start at building tools for machine learning developers and policymakers that help them achieve fairness and equity in enacting policies. In addition to this framework, we also need to develop materials and workshops for both of those audiences to help them understand the impact of these biases and to make informed policy decisions in the presence of AI-based decision making.

\bibliographystyle{unsrt}

\bibliography{refs}
\newpage
\newpage
\appendix

\section{Reproducible Audit}

To show the utility of Aequitas, we used it to audit ML models developed at the Center for Data Science and Public Policy, University of Chicago, being used to solve problems in different domains. However, we are not allowed to share the input data used in the audits. Here we present a short case study of one of our audits using a publicly available data set (COMPAS) from criminal justice. A Jupyter notebook containing the code to run this audit is available at the Aequitas Github repo \footnote{\url{https://bit.ly/2MP89VS}}.

The goal of COMPAS was to identify individuals who are at risk of recidivism to support pretrial release decisions. In a recent widely popularized investigation conducted by a team at ProPublica, Angwin et al. concluded that it was biased against black defendants. 

\subsection{Data}
The data is based on the Broward County data made publicly available by ProPublica \citep{angwin2016machine}. This data set contains COMPAS recidivism risk decile scores, 2-year recidivism
outcomes, and a number of demographic and variables on 7214 individuals who were scored
in 2013 and 2014. 

\subsection{Bias Audit using Aequitas}
We audited the predictions using Aequitas. We find that there is indeed unfairness (both unsupervised and supervised) in the model in all three attributes of interest: age, gender, and race. 

\begin{figure*}[h]
	\centering
	\includegraphics[width=1.0\linewidth]{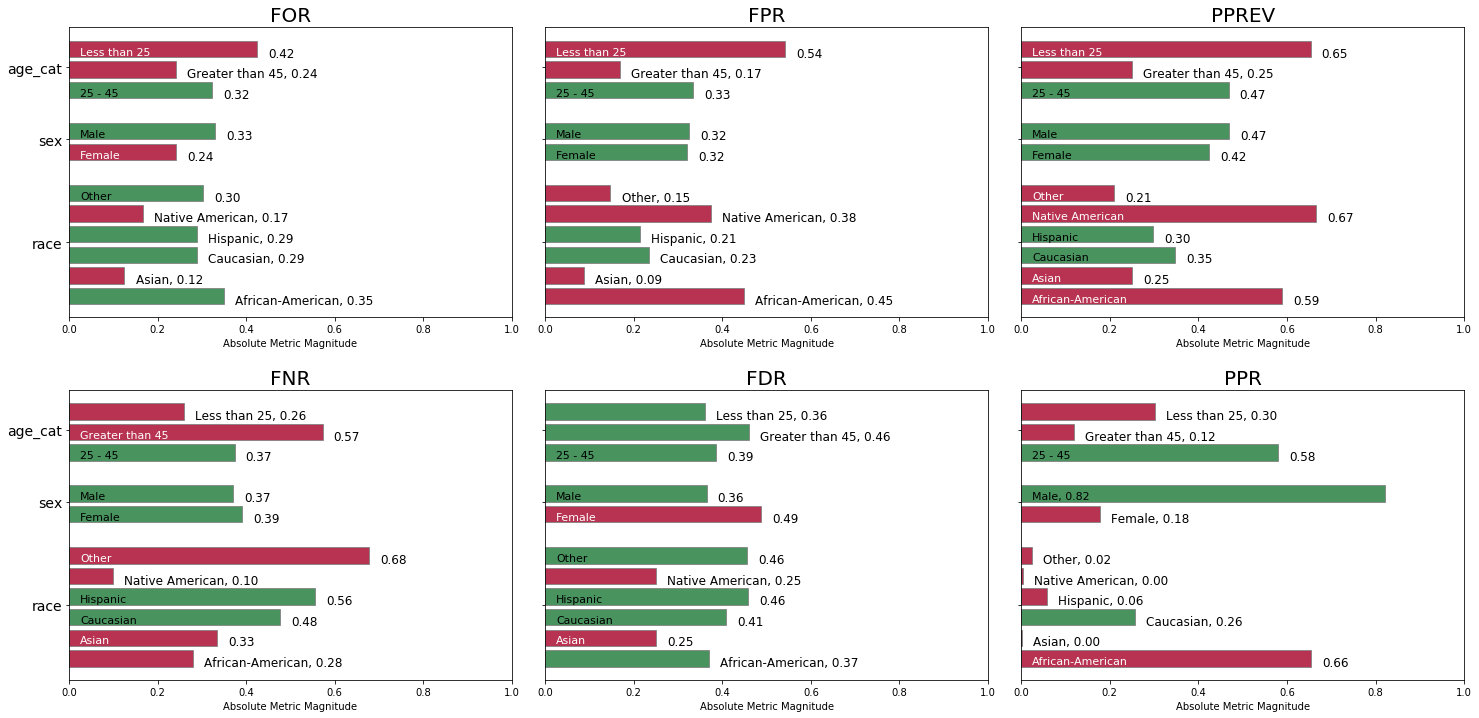}
	\caption{Group metrics results for COMPAS. Bars in green/red represent groups that passed/failed the fairness test using $\uptau=0.8$. }
	\label{fig:group_metrics}
\end{figure*}

Figure \ref{fig:group_metrics} shows the detailed group metric results. Each row represents a specific attribute-value pair (gender:female), and each bar (column) represents a group metric of interest (False Positive Rate for example). Green bars represent groups for which the model does not exhibit bias within that metric. Red bars are those that are unfavorably biased compared to the reference group. In this case we used a fairness threshold $\uptau = 0.8$ and the results show that for every metric considered there is some kind of bias towards specific groups. For instance, PPR results show that COMPAS mostly consider as high risk people with age 18-25, Males and African-Americans and that compared to each group size, younger people, Native Americans and African-Americans are being selected disproportionally.

\begin{figure*}[h]
	\centering
	\includegraphics[width=1.0\linewidth]{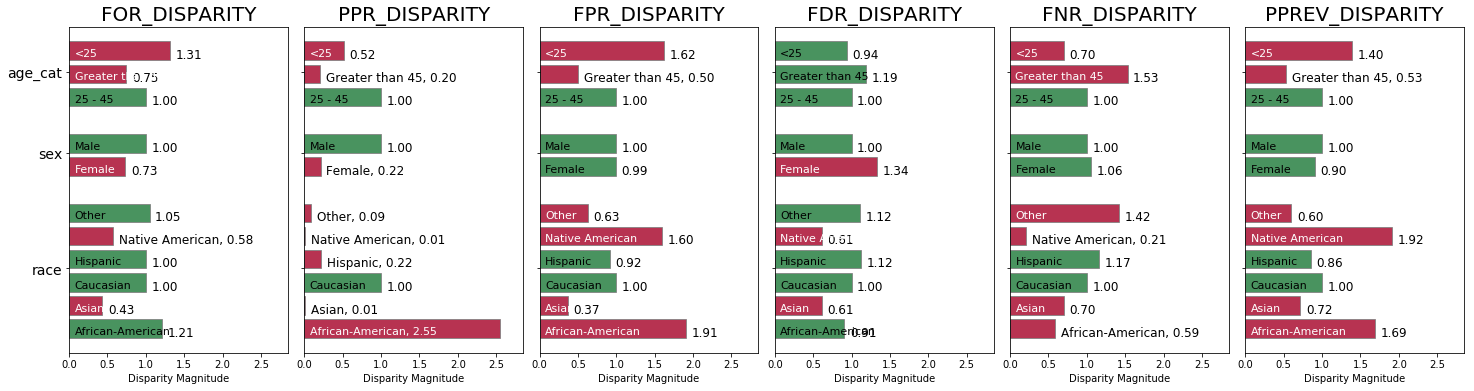}
	\caption{Disparity metrics results for COMPAS. Bars in green/red represent groups that passed/failed the fairness test using $\uptau=0.8$. }
	\label{fig:disparity}
\end{figure*}

To easily visualize the disparities between the different groups, the tool also produces results for the bias measures. We then have to determine which bias measure is relevant for our setting. If the interventions we're focusing on in our setting are ``assistive'', we only need to consider Type II parity -- False Omission Rates (FOR) and False Negative Rates (FNR). This is because our interventions are preventative and designed to provide extra assistance to individuals. Providing this assistance to individuals who are false positives will not hurt them but missing individuals could be harmful to them.

If the interventions we're focusing on in our setting are ``punitive'', we  need to consider Type I parity -- False Discovery Rates (FDR) and False Positive Rates (FPR). This is because our interventions are punitive providing this intervention to individuals who are false positives will hurt them. Since in the COMPAS setting, the predictions are being used to make pretrial release decisions, we care about FPR and FDR Parity.

Looking at the figure \ref{fig:disparity} we can see that COMPAS is fair regarding the FDR for race but as ProPublica found, the FPR for African-Americans is almost twice as the FPR for Caucasians. For age we observe the same results for the same two metrics: FDR results are fair but FPR for <25 is 1.6X higher than 25-45. On the other hand, if we consider false positive errors distribution considering Sex we observe the contrary: the model is fair for FPR but the FDR of Female is 1.34 times higher than for Male.

\end{document}